\documentclass[runningheads]{llncs}

\usepackage{eccv}

\usepackage{eccvabbrv}

\usepackage{graphicx}
\usepackage{subcaption}
\usepackage{booktabs}
\usepackage{multirow}
\usepackage{array}
\usepackage{amsmath}
\usepackage{marvosym}
\usepackage[table]{xcolor}
\usepackage{pifont}

\newcommand{\cmark}{\textcolor{green!60!black}{\ding{51}}} 
\newcommand{\xmark}{\textcolor{red}{\ding{55}}} 
\newcommand{\pmark}{\textcolor{gray}{$\sim$}} 

\definecolor{HeaderGray}{RGB}{235, 235, 240} 

\definecolor{LightBlueRow}{RGB}{248, 252, 255} 

\definecolor{OursHighlight}{RGB}{245, 240, 255} 

\usepackage[accsupp]{axessibility}  

\usepackage[pagebackref,breaklinks,colorlinks,citecolor=eccvblue]{hyperref}

\usepackage{orcidlink}
\usepackage{wrapfig}

\begin{document}

\title{PhyEditBench: A Real-World Multi-Stage Benchmark for Physics-Aware Image Editing}

\titlerunning{PhyEditBench}


\author{Shengbin Guo$^{*,1}$\and Shaokang He$^{*,2}$\and Chaoyue Meng$^{3}$\and Shengpeng Xiao$^{4}$\and \\ Xunzhi Xiang$^{1}$\and Shaofeng Zhang$^{5}$\and Qi Fan$^{1, \text{\Letter}}$}
\authorrunning{S.~Guo et al.}

\institute{
$^{1}$ Nanjing University \qquad
$^{2}$ SDU \qquad
$^{3}$ HRBEU \qquad
$^{4}$ SDUTCM \qquad
$^{5}$ USTC \\
\email{shengbinguo2022@gmail.com, sk\_he@mail.sdu.edu.cn, fanqi@nju.edu.cn}
}

\maketitle

\renewcommand{\thefootnote}{\ensuremath{*}}
\footnotetext[1]{Equal contribution, \textsuperscript{\Letter} Corresponding author.}
\renewcommand*{\thefootnote}{\arabic{footnote}}

\begin{figure}[htbp]
    \centering
    \includegraphics[width=0.93\linewidth]{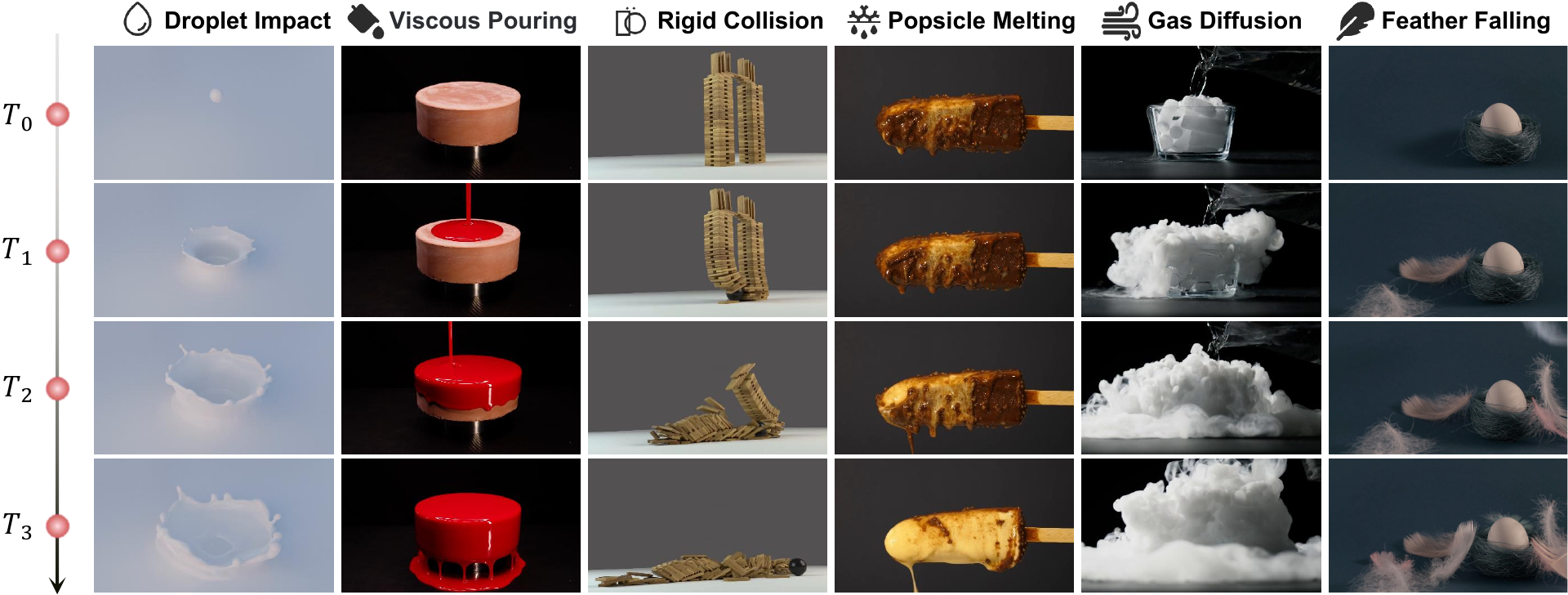}
    \caption{High-quality, high-resolution, real-world examples from PhyEditBench, which encompasses a diverse range of complex physical processes.}
    \label{fig:example}
\end{figure}

\begin{abstract}
While instruction-based image editing, enabled by multi-modal generative models, has advanced significantly, existing benchmarks lack comprehensive evaluation of physics-based reasoning—a critical capability for handling real-world scenarios. To address this, we introduce PhyEditBench, a benchmark designed to assess the physical understanding of editing models. Guided by a hierarchical taxonomy, we establish 4 primary classes and 12 subclasses. It comprises 238 \textbf{high-quality, high-resolution, real-world} instances—meticulously extracted from videos to capture authentic physical dynamics, alongside 35 synthetic Anti-Physics instances. Our empirical analysis of current SOTA editing methods exposes substantial limitations in their physics-based reasoning. We further propose a training-free baseline named \textbf{PhyWorld} that uses test-time scaling and a latent reduction strategy. PhyWorld outperforms comparable models and suggests that the video generation process can effectively serve as a reasoning mechanism for image editing. The project page is available at \url{https://github.com/Previsior/PhyEditBench}.
\end{abstract}

\section{Introduction}\label{sec:intro}

The rapid evolution of multi-modal generative models has driven unprecedented breakthroughs in both image and video generation \cite{LDM, DALL-E, sora, wan2025, Stable-video-diffusion}. Building upon these powerful foundational models, instruction-based image editing has emerged as a crucial and highly practical downstream task. It enables users to manipulate specific visual content using natural language prompts \cite{magicbrush, instructpix2pix, prompt-to-prompt, gpt, gemini, seedream, omnigen2, flux, step1x, qwen, uniworld, bagel, f2f}. Unlike generating content from scratch, the editing task requires a delicate equilibrium: models must accurately execute the given instructions while meticulously preserving the structural integrity and task-irrelevant semantics of the source image \cite{imagic, text2live, null-text}.

Early instruction-based editing models primarily focused on low-level visual transformations, such as global style transfer, color adjustment, or simple object replacement \cite{instructpix2pix, prompt-to-prompt, sdedit}. However, as user demands grow more sophisticated, there is an increasing need for editing models capable of handling complex instructions that require deep cognitive reasoning. To systematically assess and improve these capabilities, recent research has pivoted towards reasoning-centric image editing. A variety of reasoning benchmarks~\cite{krisbench, risebench, unireditbench, wiseedit} and editing frameworks have been introduced \cite{thinkrl, thinkgen, editthinker, reasonedit, diffthinker, uni-cot, mmada, cogniedit, imagent, if-edit}, pushing the boundaries of image editing from pixel manipulation to semantic-level deduction.

Despite these advancements, existing reasoning benchmarks exhibit a critical limitation: they predominantly focus on spatial layout, logic puzzles, or attribute binding \cite{krisbench, risebench, unireditbench, wiseedit, geneval, wu2025chronoedit, imgedit}. In essence, they often reduce the evaluation of ``reasoning'' to advanced instruction following and visual perception, but critically lack scenarios involving real-world physical dynamics. Consequently, these benchmarks fall short in accurately evaluating a model's intrinsic understanding of the real world. A robust editing model designed for real-world applications must not only understand \textit{what} to change semantically but also \textit{how} physical laws govern those state transitions.

To bridge this gap, we introduce \textbf{PhyEditBench}, a comprehensive and challenging benchmark specifically designed to assess the physics-based reasoning capabilities of image editing models. As shown in \cref{fig:example}, PhyEditBench is meticulously constructed from high-quality, high-resolution real-world videos to capture authentic physical dynamics. We conduct an extensive empirical analysis of current state-of-the-art (SOTA) open-source and closed-source editing models on our benchmark. The evaluation reveals substantial limitations: traditional editing methods perform sub-optimally, frequently generating physically implausible artifacts or pasting objects without logical state transitions. This indicates that current image editing paradigms, which heavily rely on static statistical priors, struggle to reason about the dynamic evolution of the physical world.

Recently, the remarkable progress in world models and video generation has offered a promising new perspective \cite{wan2025, sora, structure, videopoet}. Since video models are trained to predict subsequent frames, they inherently learn to simulate physical laws and maintain temporal causality. Inspired by this, we present a training-free framework named \textbf{PhyWorld} that leverages pretrained video generation models for physics-aware image editing. Following recent paradigms \cite{f2f, wu2025chronoedit}, we formulate the editing task as a temporal transformation process and interpret the intermediate generated frames as an implicit reasoning process. Furthermore, drawing inspiration from scaling laws at inference time \cite{he2025scaling}, we construct our method upon an evolutionary Test-Time Scaling (TTS) algorithm and a Video Reward Model \cite{video_reward_model} to iteratively optimize and substantially enhance the output quality. By integrating a latent reduction strategy, our framework ensures strict fidelity to the source image while achieving physically plausible editing results.

In summary, our main contributions are three-fold:
\begin{itemize}
    \item We introduce \textbf{PhyEditBench}, a novel, high-quality, real-world benchmark dedicated to evaluating physical reasoning in multi-modal image editing.
    \item We provide a comprehensive empirical analysis of current SOTA editing models, exposing their critical limitations in understanding and executing real-world physical dynamics.
    \item We propose a training-free editing baseline named \textbf{PhyWorld} that harnesses video generation models as reasoning engines. Augmented with Test-Time Scaling and a latent reduction strategy, our method outperforms comparable models in physically grounded editing tasks.
\end{itemize}
\section{Related Works}\label{sec:related}
\paragraph{Instruction-based Image Editing and Benchmarks.}
The rapid development of diffusion models \cite{dit, LDM}
and Multi-modal Large Language Models (MLLMs) \cite{llava, gpt4, gemini-tech} has catalyzed a paradigm shift in image editing. Early instruction-based editing methods, such as InstructPix2Pix \cite{instructpix2pix} and MagicBrush \cite{magicbrush}, primarily focused on aligning text prompts with visual representations to perform low-level manipulations, including global style transfer, color adjustment, and simple object replacement. However, as user instructions become increasingly complex, these models often struggle to comprehend the underlying logic, relying instead on superficial semantic matching.

To address this, recent research has pivoted toward reasoning-centric image editing, empowering models with deep cognitive abilities to interpret complex, multi-step, or counterfactual instructions \cite{smartedit, controlthinker, editthinker}.
Consequently, a variety of benchmarks have been proposed to systematically evaluate these advanced capabilities. For instance, KRIS-Bench \cite{krisbench} and RISEBench \cite{risebench} evaluate models across diverse cognitive dimensions, including spatial reasoning, conceptual knowledge, and logic puzzles. Similarly, UniREditBench~\cite{unireditbench} and WiseEdit~\cite{wiseedit} introduce a unified evaluation framework for reasoning-based editing. Despite their comprehensive coverage of spatial and semantic logic, these benchmarks exhibit a critical blind spot: they inherently lack the evaluation of \textit{real-world physical dynamics}. They predominantly test whether a model knows \textit{what} to change (e.g., modifying attributes or layouts) rather than \textit{how} an object's state evolves under physical laws (e.g., gravity, fluid dynamics, or deformation). Our proposed \textbf{PhyEditBench} directly addresses this gap by introducing high-resolution, real-world instances dedicated exclusively to physics-based reasoning.

\definecolor{HeaderGray}{RGB}{235, 235, 240}
\definecolor{LightBlueRow}{RGB}{243, 252, 255}
\definecolor{OursHighlight}{RGB}{245, 240, 255}
\begin{table*}[t]
\caption{Comparison between our PhyEditBench and previous related datasets. Existing image editing benchmarks lack real-world physical dynamics and multi-state granularity, while physical video datasets are not tailored for instruction-based image editing. PhyEditBench bridges this gap. (\cmark: Yes, \pmark: Partial, \xmark: No)}
\label{tab:dataset_comparison}
\centering
\resizebox{\textwidth}{!}{
\begin{tabular}{l l | c c c c c}
\toprule
\rowcolor{HeaderGray}

\textbf{Category} & \textbf{Dataset} & \textbf{Editing Task} & \textbf{Real-world} & \textbf{High-Res.} & \textbf{Physics-aware} & \textbf{Multi-State} \\
\midrule
\multirow{4}{*}{\begin{tabular}[c]{@{}l@{}}Image Editing\\ Benchmarks\end{tabular}} 
 &  RISEBench \cite{risebench}     & \cmark & \xmark & \cmark & \xmark & \xmark \\
 & KRIS-Bench \cite{krisbench}    & \cmark & \pmark & \cmark & \pmark & \xmark \\
 &  UniREditBench \cite{unireditbench}  & \cmark & \pmark & \cmark & \xmark & \xmark \\
 & WiseEdit \cite{wiseedit}       & \cmark & \pmark & \cmark & \xmark & \xmark \\
\midrule
\multirow{5}{*}{\begin{tabular}[c]{@{}l@{}}Video \& Physics\\ Datasets\end{tabular}} 
 &  CLEVRER \cite{clevrer}         & \xmark & \xmark & \xmark & \cmark & \cmark \\
 & Physion \cite{physion}         & \xmark & \xmark & \pmark & \cmark & \cmark \\
 &  NewtonGen \cite{newtongen}     & \xmark & \xmark & \cmark & \cmark & \cmark \\
 & Action100M \cite{action100m}   & \xmark & \cmark & \pmark & \xmark & \cmark \\
 &  Sth-Sth-V2 \cite{ssv2}         & \xmark & \cmark & \xmark & \cmark & \cmark \\
\midrule
\rowcolor{OursHighlight} \textbf{Ours} & \textbf{PhyEditBench} & \cmark & \cmark & \cmark & \cmark & \cmark \\
\bottomrule
\end{tabular}
}
\end{table*}

\paragraph{Physical Reasoning in Vision Models.}
Understanding intuitive physics is a fundamental hallmark of machine intelligence. Early explorations in physical reasoning primarily focused on Visual Question Answering (VQA) and video prediction tasks. Datasets such as CLEVRER~\cite{clevrer} and Physion~\cite{physion} evaluate a model's ability to predict collisions, stability, and dynamic events. While foundational, these datasets are overwhelmingly constructed using 3D rendering engines (e.g., Blender, MuJoCo), resulting in simplified, synthetic environments that fail to capture the complexity, textures, and unpredictable nature of the real world.

Recently, the intersection of physical reasoning and generative AI has garnered significant attention. Researchers have observed that despite generating visually stunning images and videos, modern diffusion models frequently violate basic physical principles, producing hallucinations such as reversed gravity or unnatural fluid dynamics \cite{phyvllm, physdreamer}.
To mitigate this, pioneering works like NewtonGen~\cite{newtongen} and PhyGDPO~\cite{phygdpo} have attempted to explicitly inject Newtonian dynamics or physics-guided reward models into the generation process. However, these efforts are largely confined to text-to-video generation or rely heavily on external physical simulators. To date, there is a conspicuous absence of a benchmark designed to evaluate how well \textit{general image editing models} understand and manipulate real-world physical states. As compared in \cref{tab:dataset_comparison}, existing resources fall into two disjoint extremes: current editing benchmarks lack physical and multi-state granularity, while physical video datasets are ill-suited for instruction-guided editing tasks. By curating a taxonomy of real-world physical interactions, our benchmark serves as a crucial testbed to bridge this gap.

\paragraph{World Models and Video Generation.}
The recent emergence of large-scale video generation models, often referred to as ``world models'' (e.g., Sora~\cite{sora}, Stable Video Diffusion \cite{wan2025, Stable-video-diffusion}), has demonstrated unprecedented capabilities in simulating the physical world. By training on massive amounts of sequential data to predict subsequent frames, these models implicitly internalize physical laws, temporal causality, and object persistence \cite{wan2025, DALL-E, videopoet}.

This profound temporal and physical understanding presents a novel pathway for solving complex image editing tasks. Instead of treating editing as a static, single-step pixel transformation, recent state-of-the-art approaches have begun to formulate image editing as a temporal generation process. Methods such as CoF~\cite{cof}, Frame2Frame~\cite{f2f}, and ChronoEdit~\cite{wu2025chronoedit} leverage pretrained video diffusion models to generate intermediate transition frames, effectively utilizing the video generation process as an implicit reasoning mechanism. 
Inspired by this paradigm, we leverage world models to address physics-based editing. Specifically, our training-free framework employs a test-time optimization approach to enhance generation quality. By transforming static physics editing instructions into a video-guided reasoning trajectory, our method harnesses the physically plausible reasoning capabilities inherent in pretrained video generation models. Consequently, it outperforms most traditional static editing models on physically demanding tasks, despite maintaining a compact 5B parameter size.
\section{PhyEditBench}\label{sec:bench}
\subsection{Overview}\label{sec:bench_overview}

PhyEditBench is a benchmark for evaluating physics-based reasoning in instruction-guided image editing. 
Unlike existing benchmarks that primarily emphasize semantic correctness or local appearance edits, PhyEditBench targets \emph{physical process understanding}: models must produce visually faithful edits that follow plausible physical dynamics and maintain scene invariants.

\begin{figure}[t]
    \centering
    \begin{subfigure}[b]{0.49\linewidth}
        \centering
        \includegraphics[width=\linewidth]{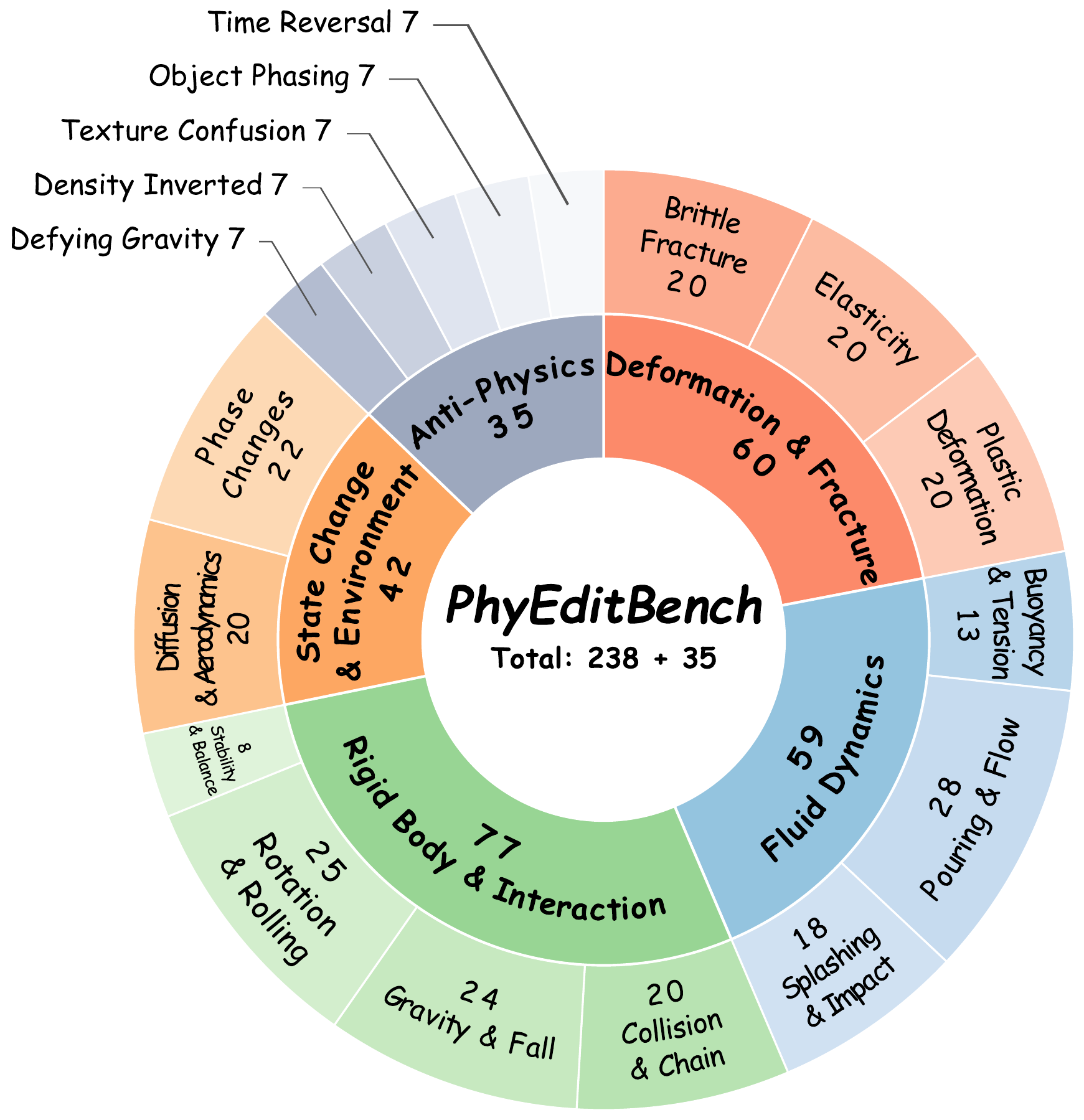}
        \caption{Benchmark Overview}
    \end{subfigure}
    \hfill
    \begin{subfigure}[b]{0.49\linewidth}
        \centering
        \includegraphics[width=\linewidth]{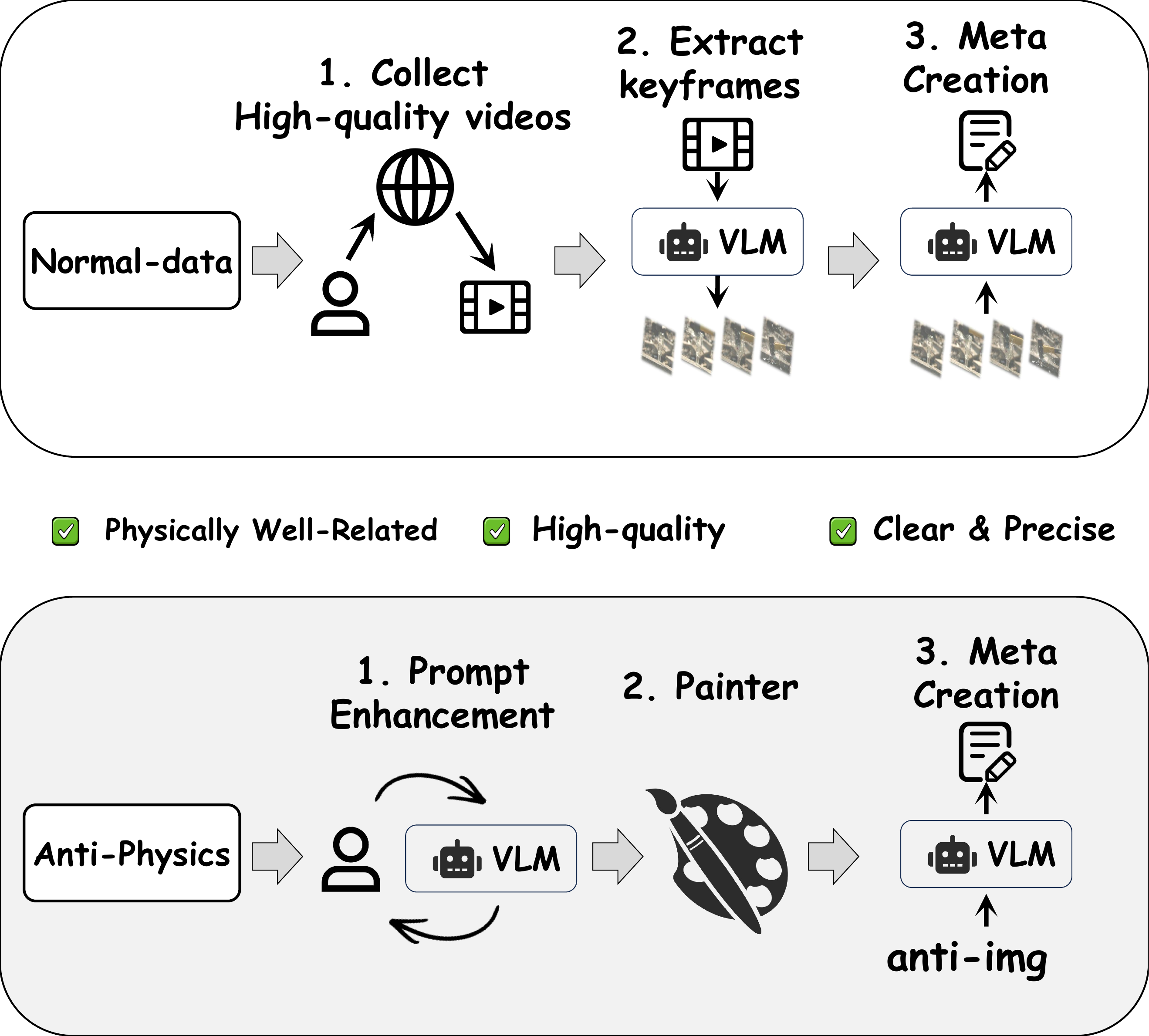}
        \caption{Data Construction}
    \end{subfigure}
    \caption{\textbf{(a)} shows our benchmark taxonomy and data volume. \textbf{(b)} illustrates the data construction pipeline.}
    \label{fig:overview}
\end{figure}

\paragraph{Benchmark composition.}
Guided by a hierarchical taxonomy, PhyEditBench contains 4 primary classes and 12 subclasses spanning common real-world physical phenomena. 
The benchmark includes 238 instances extracted from real-world videos to capture authentic physical dynamics, and an additional 35 synthetic \emph{Anti-Physics} instances that deliberately violate physical laws.
\cref{fig:overview}~(a) provides an overview of the taxonomy and data distribution.

\begin{figure}[t]
    \centering
    \includegraphics[width=1\linewidth]{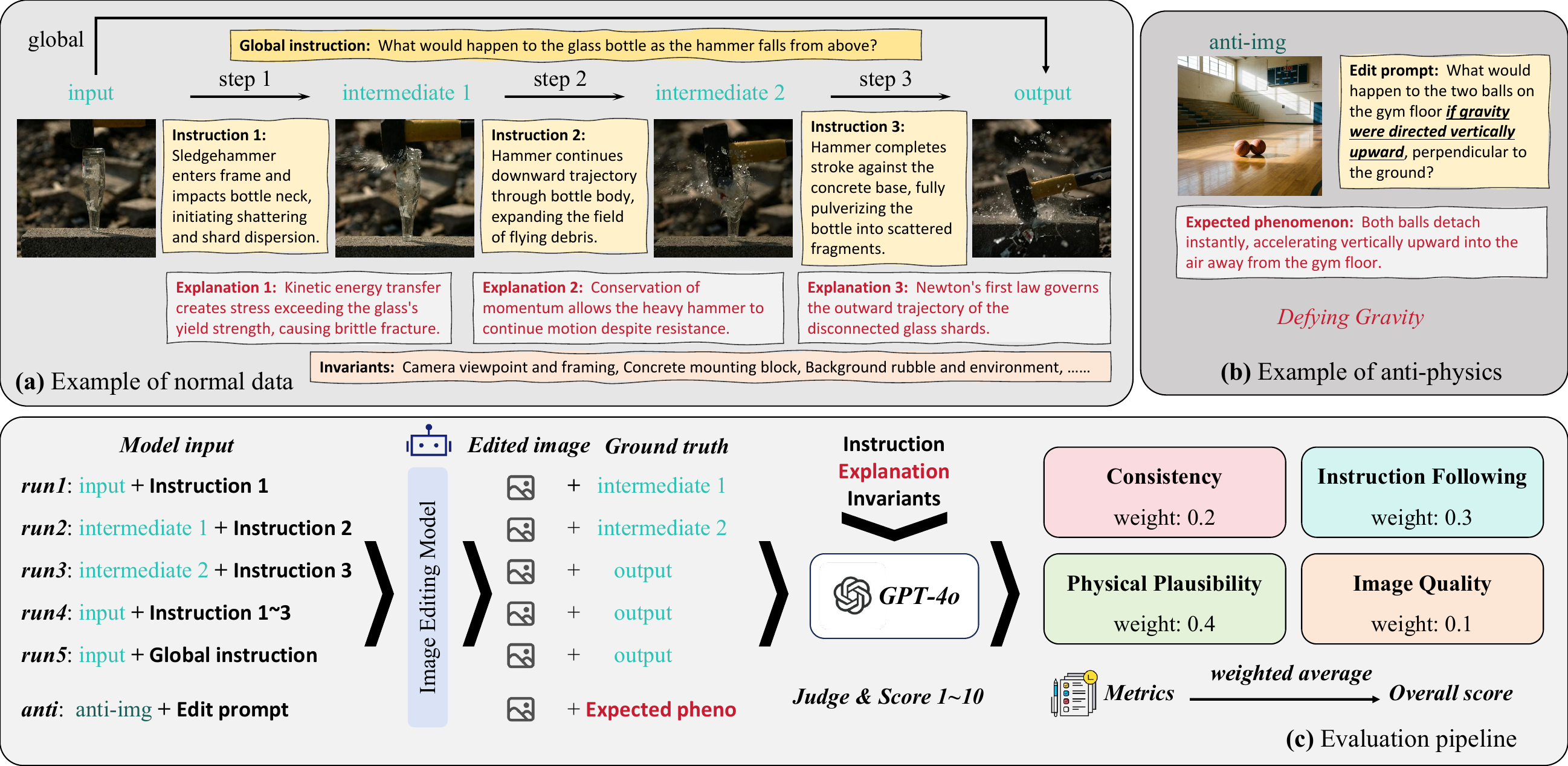}
    \caption{\textbf{(a)} shows the form of normal data points, including pictures, editing instructions, explanations, and invariants. \textbf{(b)} depicts the data point form of anti-physics, including original images, editing instructions that violate physics, and expected phenomena. \textbf{(c)} illustrates our benchmark scoring pipeline.}
    \label{fig:data_eval}
\end{figure}

\paragraph{Instance format.}
\cref{fig:data_eval}~(a) shows the composition of normal data points.
Each instance in the physics-process subset is represented as a \emph{four-state trajectory}:
\texttt{input}, \texttt{intermediate 1}, \texttt{intermediate 2}, and \texttt{output}. 
These four frames are sampled from a single real video to depict a temporally coherent physical transition from an initial stable state to a final state.
To evaluate both coarse and fine-grained physical understanding, each instance is annotated with:
(i) one \textbf{global instruction} describing the overall edit goal from \texttt{input} to \texttt{output};
(ii) three \textbf{step instructions} describing the intended transition for each consecutive pair of states; 
(iii) concise \textbf{explanations} of the underlying physical process; and
(iv) \textbf{invariants} (e.g., viewpoint and background) that should remain unchanged.
This design supports two complementary evaluation settings: 
(i) \emph{step-wise editing}, which tests whether a model can follow physically meaningful intermediate checkpoints; and 
(ii) \emph{global editing}, which tests whether the model can infer plausible intermediate dynamics from a high-level instruction.

\paragraph{Why intermediate states?}
Physical processes often unfold gradually and contain latent constraints that are not captured by a single end-state comparison~\cite{cot, visualcot, sora, Stable-video-diffusion}. 
Intermediate checkpoints in PhyEditBench enable fine-grained diagnosis: a model may reach the final state while violating the physical trajectory (e.g., implausible motion direction or inconsistent material evolution), which would be exposed by mismatches at \texttt{intermediate 1}/\texttt{intermediate 2}. 
Consequently, this multi-stage structure provides a significantly stronger and more rigorous probe for physics-grounded reasoning compared to standard one-shot edits.

\subsection{Taxonomy of Physical Types}\label{sec:taxonomy}
To systematically cover different real-world physics while maintaining the interpretability of the benchmark, we designed a hierarchical taxonomy.
Inspired by previous works \cite{physion, clevrer, spelke1990principles}, each subclass is defined by a set of physical principles and representative scene patterns, enabling both category-level analysis and fine-grained diagnosis.

\paragraph{Deformation \& Fracture.}
This primary class captures shape change and material failure under external forces, emphasizing how object geometry and integrity evolve with impact, compression, and elastic recovery. 
It includes three subclasses: \emph{Brittle Fracture}, \emph{Plastic Deformation}, and \emph{Elasticity}.

\paragraph{Fluid Dynamics.}
This class focuses on complex liquid motion and multi-phase phenomena, where realistic edits require coherent free-surface behavior, plausible flow patterns, and physically consistent interactions between fluids and interacting objects. 
To systematically cover these dynamics, it includes three distinct subclasses: \emph{Splashing \& Impact}, \emph{Pouring \& Flow}, and \emph{Buoyancy \& Tension}.

\paragraph{Rigid Body \& Interaction.}
This class covers rigid-body motion and contact-driven interactions governed by gravity, momentum transfer, friction, stability, and rotation. 
It includes four subclasses: \emph{Gravity \& Fall}, \emph{Collision \& Chain}, \emph{Stability \& Balance}, and \emph{Rotation \& Rolling}.

\paragraph{State Change \& Environment.}
This class describes transformations induced by environmental factors such as heat transfer and air/gas dynamics, which often manifest as gradual changes with characteristic visual signatures. 
It includes two subclasses: \emph{Phase Changes} and \emph{Diffusion \& Aerodynamics}.

\paragraph{Anti-Physics Data point.}
As illustrated in \cref{fig:data_eval}~(b), an Anti-Physics data point is meticulously designed to evaluate a model's ability to process counterfactual physical conditions. Each instance comprises three components: a source image, an edit prompt, and an expected phenomenon. The edit prompt explicitly injects a physical rule that contradicts common real-world experience.
The inclusion of Anti-Physics instances is essential to decouple genuine physical reasoning from memorized statistical priors. Conventional models often rely on pre-training biases (e.g., assuming ``knives always cut apples'') and ignore explicit textual constraints \cite{clevrer, marcus2020next, winoground, shortcut}. By introducing counterfactual scenarios, we force models to suppress these inherent visual habits and execute strict deductive reasoning based solely on the prompt. Success in these instances thus demonstrates true dynamic physical deduction rather than mere pattern matching.

\subsection{Data Construction}\label{sec:construction}
The overall pipeline is shown in \cref{fig:overview}~(b). For real-world instances, we sourced royalty-free videos from public stock platforms. We utilized a Vision-Language Model for coarse keyframe proposal and metadata drafting, followed by rigorous human verification and correction of temporal orders, keyframe alignment, and physical invariants to ensure high quality. For Anti-Physics data, the source images were synthesized using modern generative models, with the VLM formulating the counterfactual prompts and expected phenomena, followed by human auditing. Details are provided in the Appendix.

\subsection{Evaluation Pipeline}\label{sec:eval_pipe}

Following previous work \cite{risebench, krisbench, wiseedit, unireditbench}, PhyEditBench evaluates editing models by generating edited images under standardized inputs and then scoring the results by four dimensions with a unified VLM-based judge (GPT-4o). \cref{fig:data_eval}~(c) illustrates the overall evaluation pipeline.

\paragraph{Model inputs and run types.}
For the physics-process subset, each instance contains four states (\texttt{input}, \texttt{intermediate 1}, \texttt{intermediate 2}, \texttt{output}), and we define five run types to probe both fine-grained and holistic understanding. 
Runs \textbf{TypeA--TypeC} perform step-wise editing on consecutive state pairs: \texttt{input}\,$\rightarrow$\,\texttt{intermediate 1}, \texttt{intermediate 1}\,$\rightarrow$\,\texttt{intermediate 2}, and \texttt{intermed\-iate 2}\,$\rightarrow$\,\texttt{output}, each taking the corresponding input image and step instruction to produce one edited image. 
\textbf{TypeD} applies the three-step instructions jointly starting from \texttt{input} and evaluates only the final state. 
\textbf{TypeE} performs global editing from \texttt{input} to \texttt{output} using the high-level instruction.
For the Anti-Physics subset, the editing model takes a single input image and a counterfactual edit prompt, and outputs one edited image.

\paragraph{VLM-based scoring.}
Given the model output, GPT-4o scores each run type using the provided instruction or edit prompt, optional physical explanation, and invariants, together with the relevant reference images (ground-truth targets when available). 
Specifically, GPT-4o assigns a score in \{1,\dots,10\} with a brief rationale for four complementary dimensions:
\textbf{1. Consistency}, preservation of invariants and non-target content.
\textbf{2. Instruction Following}, faithfulness to the instruction, and alignment with the intended target state.
\textbf{3. Physical Plausibility}, whether the edit reflects physically coherent dynamics consistent with the provided explanation or expected phenomenon.
\textbf{4. Image Quality}, visual realism, and absence of artifacts.
We compute the final score as a weighted average of these four dimensions, using fixed weights across all runs and both subsets. Details are provided in the Appendix.

\section{Method}\label{sec:method}
\begin{figure}[t]
    \centering
    \includegraphics[width=1\linewidth]{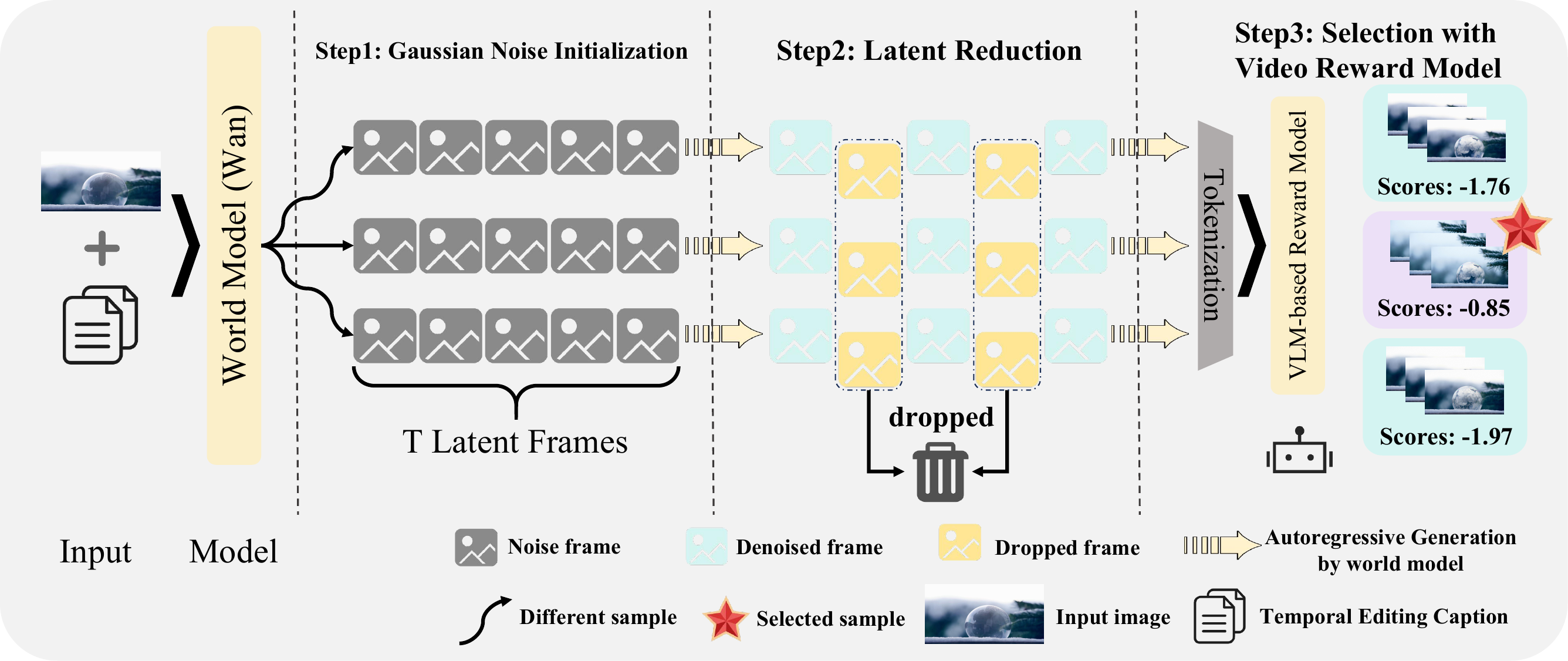}
    \caption{Overview of the proposed PhyWorld pipeline. The editing process begins by initializing multiple Gaussian noise samples. During generation, a latent reduction strategy dynamically drops intermediate frames to compress the sequence and improve efficiency. Finally, a Video Reward Model evaluates all generated candidates, selecting the optimal sequence whose final frame is then extracted as the editing result.}
    \label{fig:pipeline}
\end{figure}

\subsection{Overview} This work introduces \textbf{PhyWorld}, a strong \textbf{training-free} baseline that leverages pretrained video generation models for image editing. Aligning with~\cite{f2f, wu2025chronoedit}, we formulate editing as a temporal transformation, interpreting intermediate frames as a reasoning process similar to ChronoEdit~\cite{wu2025chronoedit}. Inspired by~\cite{he2025scaling}, we build PhyWorld upon an evolutionary Test-Time Scaling (TTS) algorithm and a Video Reward Model~\cite{video_reward_model} to elevate output quality, while incorporating a latent reduction strategy to ensure efficiency. Our method achieves superior performance on our benchmark, outperforming same-category methods with comparable parameter sizes~\cite{f2f} as well as most existing open-source models. The overall architecture is depicted in~\cref{fig:pipeline}.

\subsection{Editing Prompt Enhancement} Conventionally, text-based image editing methods operate on an image-text pair $(I_s, c)$, where the text $c$ serves as guidance for the specific modifications to be applied to $I_s$. Similar to Frame2Frame~\cite{f2f}, our framework transforms the static editing prompt into a format suitable for video generation. Specifically, we adapt the original instruction into a \textit{Temporal Editing Caption}~\cite{f2f}, designed to describe the transition process between the input and output images. Leveraging recent advances in vision-language models (VLMs), we employ Qwen-3.5 Max~\cite{qwen35blog}, a state-of-the-art VLM, to perform this enhancement. In particular, the model reasons about the actions described in the editing prompt, analyzes their physical procedure, and extends them into a detailed description of the underlying physical process. The prompt is shown in the Appendix.

\subsection{Test-Time Scaling} To enhance video generation quality, EvoSearch~\cite{he2025scaling} proposes an evolutionary search algorithm for Text-to-Video tasks that identifies the optimal output. The search process is conducted via latent selection at specific denoising timesteps using a video reward model~\cite{video_reward_model}, formulated as:
\begin{equation}
R(\boldsymbol{x}_{t_i}) = \mathbb{E}_{\boldsymbol{x}_0 \sim p_0(\boldsymbol{x}_0|\boldsymbol{x}_{t_i})} \left[ r(\boldsymbol{x}_0) | \boldsymbol{x}_{t_i} \right],
\label{eq:reward}
\end{equation}\\
The method first randomly initializes a population of latents $\{\boldsymbol{x}_T^i\}_{i=1}^{k_{\text{start}}}$ at timestep $T$, where $k_{\text{start}}$ denotes the initial population size in the population size schedule ~$ k=\{k_{\text{start}}, k_{t_1}, \dots, k_{t_j}, \dots\} $. At each evolution timestep $t_j$ in evolution schedule $\mathcal{T} = \{T, \dots, t_j, \dots, t_n\}$, EvoSearch performs evolutionary optimization on the current latents by denoising the latents into videos and scoring them according to the formulation above~\cref{eq:reward}. The method then stores the latents at each evolution timestep along with their corresponding video scores. Through Top-K selection, tournament selection, and mutation operations, the method selects a subset of latents to continue the subsequent denoising process. Despite its improvements in video generation quality, this approach incurs substantial computational overhead, resulting in low efficiency. Furthermore, its application is restricted to Text-to-Video generation. In this work, we adapt this method to the Image-to-Video (I2V) architecture and strike a balance between the efficacy of EvoSearch and computational cost by performing a single search step at the end of the generation process. This strategy does not significantly compromise performance while substantially improving efficiency.

\subsection{Video Generation}
We leverage the pretrained Wan2.2 video generation model~\cite{wan2025} as our backbone. Specifically, we employ its TI2V-5B variant, which utilizes the Wan2.2-VAE for efficient latent compression, achieving a spatial-temporal compression ratio of $4 \times 4 \times 16$. Formally, let $C$ denote the input instruction and $I$ denote the input image. The generation process initiates with 5 Gaussian noise samples, which undergo denoising conditioned on $I$. We utilize 121 frames as reasoning tokens and 30 sampling timesteps during generation. To optimize computational efficiency, we follow the approach in~\cite{wu2025chronoedit} by employing a method termed the latent reduction strategy, as depicted in~\cref{fig:pipeline}. Specifically, this strategy is applied at sampling timesteps 10 and 20. While the initial 121 frames are encoded into 31 VAE latent tokens, the sequence length is reduced to 21 and 11 tokens at these respective timesteps by pruning intermediate latent states. After the generation process, the best output is determined by the video reward model~\cite{video_reward_model} and the final frame of the output is selected as the edited image conditioned on $ I $ and $C$.

\begin{table*}[!t]
\caption{Main experimental results of editing models.}
\label{tab:main_results}
\centering

\newcolumntype{C}[1]{>{\centering\arraybackslash}p{#1}}

\resizebox{\linewidth}{!}{
\begin{tabular}{cc|*{3}{C{0.8cm}}|*{8}{C{0.8cm}}|*{3}{C{0.8cm}}}
\toprule
\multicolumn{2}{c|}{}
  & \multicolumn{3}{c|}{closed-source} 
  & \multicolumn{8}{c|}{open-source} 
  & \multicolumn{3}{c}{video-based} \\
\cmidrule(lr){3-5} \cmidrule(lr){6-13} \cmidrule(lr){14-16}
\rotatebox{90}{Dimension}
  & \rotatebox{90}{Model}
  & {\scriptsize\rotatebox{90}{GPT-Image-1.5}}
  & \rotatebox{90}{Gemini-2.5} 
  & \rotatebox{90}{Seedream4.0} 
  & \rotatebox{90}{OmniGen2} 
  & {\scriptsize\rotatebox{90}{InstructPix2Pix~}} 
  & {\tiny\rotatebox{90}{FLUX.1-Kontext-dev~}} 
  & \rotatebox{90}{Step1X-Edit} 
  & {\scriptsize\rotatebox{90}{Qwen-Image-Edit}} 
  & \rotatebox{90}{UniWorld-V2} 
  & \rotatebox{90}{BAGEL} 
  & \rotatebox{90}{BAGEL-Think} 
  & \rotatebox{90}{Frame2Frame} 
  & \rotatebox{90}{PhyWorld} 
  & \rotatebox{90}{ChronoEdit-14B} \\

\midrule

\rowcolor{blue!10} \multicolumn{16}{c}{\textbf{normal data}} \\
\midrule

Avg. & Overall 
    & \textcolor{gray}{8.23} & \textcolor{gray}{6.51} & \textcolor{gray}{8.47} & 5.36 & 5.61 & 5.27 & 6.79 & 6.71 & \underline{7.08} & 6.09 & 6.43 & 5.38 & 6.43 & \textbf{8.51} \\
\midrule
\multirow{4}{*}{\begin{tabular}[c]{@{}c@{}}Avg.\\ by\\ Metrics\end{tabular}} 
    & Cons.
    & \textcolor{gray}{8.84} & \textcolor{gray}{7.21} & \textcolor{gray}{9.69} & 6.35 & 7.08 & 7.03 & \underline{9.24} & 8.41 & 8.53 & 7.82 & 8.53 & 7.96 & 7.58 & \textbf{9.67} \\
    & Inst.
    & \textcolor{gray}{8.21} & \textcolor{gray}{6.13} & \textcolor{gray}{8.22} & 4.23 & 4.16 & 4.56 & 5.93 & 6.35 & \underline{6.78} & 5.47 & 5.72 & 4.08 & 5.89 & \textbf{7.98} \\
    & Phys.
    & \textcolor{gray}{8.30} & \textcolor{gray}{6.64} & \textcolor{gray}{8.22} & 5.54 & 5.58 & 4.81 & 6.19 & 6.37 & \underline{6.73} & 5.69 & 5.96 & 4.73 & 6.30 & \textbf{8.42} \\
    & Imag.
    & \textcolor{gray}{6.78} & \textcolor{gray}{5.72} & \textcolor{gray}{7.75} & 5.96 & \underline{7.12} & 5.71 & 6.87 & 5.79 & 6.45 & 6.04 & 6.26 & 6.65 & 6.26 & \textbf{8.13} \\
\midrule
\multirow{4}{*}{\begin{tabular}[c]{@{}c@{}}Avg.\\ by\\ Classes\end{tabular}} 
    & Defor.
    & \textcolor{gray}{8.36} & \textcolor{gray}{7.36} & \textcolor{gray}{7.43} & 3.71 & 4.24 & 5.40 & 3.16 & 6.41 & \underline{6.60} & 5.55 & 5.77 & 5.25 & 6.29 & \textbf{8.38} \\
    & Fluid.
    & \textcolor{gray}{8.30} & \textcolor{gray}{6.79} & \textcolor{gray}{9.04} & 6.32 & 6.41 & 5.37 & 7.20 & 7.38 & \underline{7.43} & 6.43 & 6.82 & 5.43 & 6.55 & \textbf{8.74} \\
    & Rigid.
    & \textcolor{gray}{7.91} & \textcolor{gray}{5.87} & \textcolor{gray}{8.65} & 5.51 & 5.49 & 4.99 & 6.77 & 6.10 & \underline{6.99} & 6.21 & 6.55 & 5.17 & 6.10 & \textbf{8.32} \\
    & State.
    & \textcolor{gray}{8.51} & \textcolor{gray}{6.07} & \textcolor{gray}{8.84} & 6.05 & 6.67 & 5.44 & 7.13 & 7.33 & \underline{7.44} & 6.14 & 6.59 & 5.86 & 6.94 & \textbf{8.79} \\
\midrule
\multirow{5}{*}{\begin{tabular}[c]{@{}c@{}}Avg.\\ by\\ Types\end{tabular}} 
    & TypeA 
    & \textcolor{gray}{8.16} & \textcolor{gray}{6.55} & \textcolor{gray}{8.60} & 5.44 & 5.51 & 5.53 & 7.16 & 7.01 & \underline{7.43} & 6.01 & 6.53 & 5.62 & 6.59 & \textbf{8.64} \\
    & TypeB 
    & \textcolor{gray}{8.29} & \textcolor{gray}{6.52} & \textcolor{gray}{8.60} & 5.40 & 6.23 & 5.51 & \underline{7.44} & 7.09 & 7.41 & 7.06 & 7.08 & 6.04 & 6.59 & \textbf{8.68} \\
    & TypeC 
    & \textcolor{gray}{8.35} & \textcolor{gray}{6.43} & \textcolor{gray}{8.74} & 5.30 & 5.97 & 5.36 & 7.16 & 6.83 & \underline{7.46} & 6.87 & 7.01 & 5.70 & 6.34 & \textbf{8.51} \\
    & TypeD
    & \textcolor{gray}{8.32} & \textcolor{gray}{6.68} & \textcolor{gray}{8.27} & 5.54 & 4.98 & 5.02 & 6.02 & 6.36 & \underline{6.46} & 5.49 & 5.83 & 4.35 & 6.27 & \textbf{8.49} \\
    & TypeE 
    & \textcolor{gray}{8.01} & \textcolor{gray}{6.37} & \textcolor{gray}{8.15} & 5.08 & 5.36 & 4.94 & 6.15 & 6.27 & \underline{6.64} & 4.99 & 5.72 & 5.17 & 6.36 & \textbf{8.23} \\
    
\midrule
\rowcolor{red!10} \multicolumn{16}{c}{\textbf{anti-physics}} \\
\midrule

Avg. & Overall 
    & \textcolor{gray}{7.04} & \textcolor{gray}{7.07} & \textcolor{gray}{6.95} & 4.74 & 4.45 & 6.07 & 5.34 & \underline{6.16} & 5.79 & 5.11 & \underline{6.16} & 3.81 & \textbf{6.39} & 4.99 \\
\midrule
\multirow{4}{*}{\begin{tabular}[c]{@{}c@{}}Avg.\\ by\\ Metrics\end{tabular}} 
    & Cons.
    & \textcolor{gray}{9.03} & \textcolor{gray}{8.71} & \textcolor{gray}{8.20} & 8.17 & 7.60 & \textbf{8.60} & \underline{8.23} & 7.74 & 8.17 & 7.77 & 7.89 & 6.80 & 7.91 & 6.89 \\
    & Inst.
    & \textcolor{gray}{7.20} & \textcolor{gray}{7.03} & \textcolor{gray}{7.29} & 4.09 & 3.69 & 5.71 & 4.83 & \underline{6.29} & 5.77 & 4.86 & \textbf{6.34} & 2.51 & \textbf{6.34} & 4.66 \\
    & Phys.
    & \textcolor{gray}{5.51} & \textcolor{gray}{5.91} & \textcolor{gray}{5.78} & 2.89 & 2.54 & 4.43 & 3.51 & 4.83 & 4.06 & 3.29 & \underline{5.00} & 2.51 & \textbf{5.29} & 3.63 \\
    & Imag.
    & \textcolor{gray}{8.71} & \textcolor{gray}{8.54} & \textcolor{gray}{8.11} & 7.23 & 8.09 & \textbf{8.63} & \underline{8.40} & 7.91 & 8.06 & 7.83 & 6.83 & 6.89 & 7.86 & 7.69 \\
\bottomrule
\end{tabular}
}
\end{table*}

\section{Experiments}\label{sec:experiments}

\subsection{Evaluation Models and Settings}
To evaluate representative instruction-based image editing approaches, we benchmark a diverse set of models covering both closed-source and open-source systems, as well as different generation paradigms. 
For traditional image editing, we include three closed-source models: \textbf{GPT-Image-1.5}~\cite{gpt}, \textbf{Gemini-2.5-flash-image}~\cite{gemini}, and \textbf{Seedream4.0}~\cite{seedream}, together with open-source models: \textbf{OmniGen2}~\cite{omnigen2}, \textbf{InstructPix2Pix}~\cite{instructpix2pix}, \textbf{FLUX.1-Kontext-dev}~\cite{flux}, \textbf{Step1X-Edit}~\cite{step1x}, \textbf{Qwen-Image-Edit}~\cite{qwen}, \textbf{UniWorld-V2}~\cite{uniworld}, and \textbf{BAGEL}~\cite{bagel}.
Beyond conventional editors, we additionally evaluate video-generation-based editing methods: \textbf{Frame2Frame}~\cite{f2f}, \textbf{ChronoEdit}~\cite{wu2025chronoedit}, and \textbf{PhyWorld}, which perform image editing by leveraging the video generation process as an implicit reasoning mechanism. The detail information of each model can be referred to the Appendix.

\begin{figure}[t]
    \centering
    \includegraphics[width=1\linewidth]{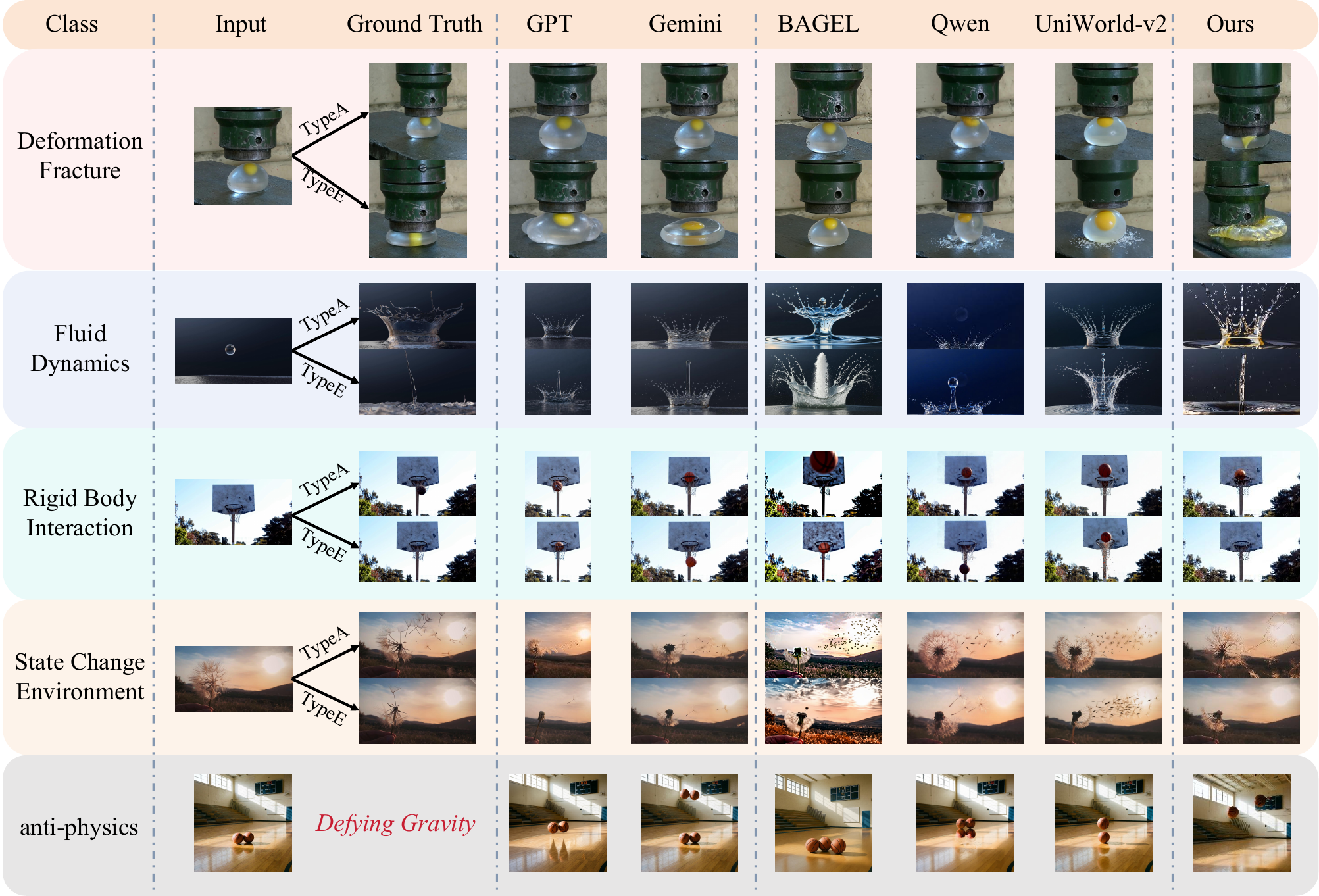}
    \caption{Qualitative comparison on physical reasoning tasks across five distinct categories. Each row presents two editing variations (Type A/E). PhyWorld demonstrates superior physical plausibility and closer alignment with ground truth, notably excelling in the anti-physics scenario.}
    \label{fig:case_study}
\end{figure}

\subsection{Main Results}
\paragraph{Overall Performance.} The empirical findings summarized in~\cref{tab:main_results} encompass performance across both normal and anti-physical data subsets, offering a comprehensive view of model capabilities. Within the conventional data setting, ChronoEdit-14B secures the top position, with Seedream4.0 trailing by a narrow margin. Turning to other open-source alternatives, UniWorld-V2 emerges as a strong performer, whereas our approach attains a highly competitive ranking—remarkably, with the most compact architecture (5B parameters vs. ~14B in ChronoEdit-14B and the similarly performing BAGEL-Think, and ~19B in Step1X-Edit-V1P1). It is also worth highlighting that BAGEL-Think surpasses its predecessor BAGEL, implying that the tasks curated in our benchmark place substantial demands on physical reasoning proficiency. 

Shifting focus to the anti-physical subset, a different picture emerges: neither closed-source nor open-source models demonstrate robust competence in tackling physically counterfactual scenarios. In this challenging regime, Gemini-2.5 maintains its leadership among proprietary systems, while PhyWorld distinguishes itself as the strongest open-source contender, closing the performance disparity relative to closed-source solutions. These findings validate that our framework effectively leverages intermediate video generation frames as reasoning tokens, while the Test-Time Scaling (TTS) strategy unlocks the pre-trained model's inherent physical reasoning capabilities. Overall, the suboptimal performance exhibited by both open-source and closed-source models on our benchmark underscores a critical limitation: current image editing models require significant advancements in reasoning about physical processes.

\paragraph{Analysis by Metrics.} Our protocol evaluates Consistency, Instruction Following, Physical Plausibility, and Image Quality. Physical Plausibility serves as the core metric for assessing real-world reasoning. In the conventional subset, GPT-Image-1.5 and ChronoEdit-14B lead the closed-source and open-source models, respectively. Notably, our training-free method secures a highly competitive place among open-source solutions. This performance is particularly encouraging considering that our framework operates without additional fine-tuning and relies entirely on a compact 5B Image-to-Video backbone architecture.

\paragraph{Analysis by Classes.} Performance varies significantly across physical categories. As shown in Tab. 2, \textit{Deformation \& Fracture} and \textit{Fluid Dynamics} are particularly challenging for most open-source static models due to complex topological changes. Conversely, closed-source models (e.g., Seedream4.0) and video-based models like ChronoEdit-14B exhibit remarkable proficiency. PhyWorld demonstrates robust, balanced performance across all categories, highlighting the versatility of video priors.

\paragraph{Analysis by Types.} Evaluating across run types reveals the compounding difficulty of long-horizon reasoning tasks. Most models achieve peak performance in short-term, single-step transitions (TypeA/B) but degrade significantly in global (TypeE) or joint multi-step (TypeD) settings. This pattern exposes the vulnerability of traditional static models to error accumulation during extended physical processes. In contrast, video-based frameworks effectively mitigate this degradation by strictly adhering to temporal causality, underscoring the inherent advantage of continuous temporal modeling for complex state transitions.

\subsection{Assessment of Evaluation Protocol}

\begin{figure}[t]
    \centering
    \includegraphics[width=1\linewidth]{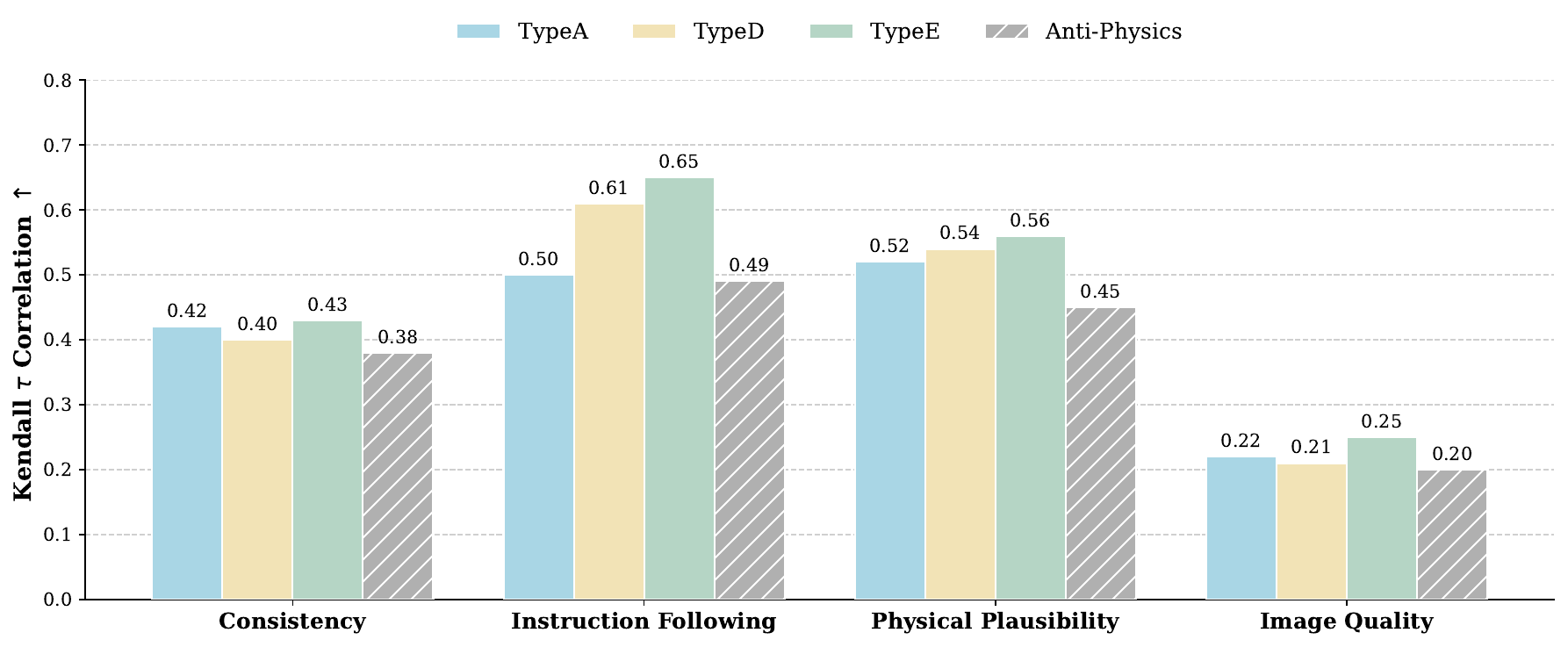}
    \caption{Correlation between human and VLM evaluations across different physical stages and metrics. We report the Kendall $\tau$ rank correlation coefficient on four metrics.}
    \label{fig:kendall}
\end{figure}

To ensure the reliability and fairness of our automated Vision-Language Model (VLM) evaluator, we conducted a human validation study. We randomly sampled instances across different temporal states of physical evolution (TypeA, TypeD, TypeE) as well as counterfactual scenarios (Anti-Physics). Human raters were provided with the source images, edit prompts, and the generated outputs from four representative models. They were instructed to rank the models across four dimensions: Visual Consistency (VC), Instruction Following (IF), Physical Plausibility (PP), and Image Quality (IQ). We then calculated the Kendall $\tau$ rank correlation coefficient to measure the alignment between human judgments and the VLM's automated rankings \cite{kendall1945treatment}. Details can be found in the Appendix.

As illustrated in \cref{fig:kendall}, the empirical results validate our evaluation protocol by demonstrating strong human-model alignment across critical reasoning dimensions. Specifically, the Kendall tau rank correlation for instruction following and physical plausibility reaches up to 0.65 and 0.56, respectively, during the final output stage, logically increasing as the physical transformations become more visually pronounced. Furthermore, the evaluator maintains robust agreement when assessing counterfactual anti-physics scenarios, proving its competence in interpreting unnatural dynamics despite the inherent subjectivity of such tasks. Conversely, the correlation for low-level image quality remains notably lower. This divergence is a well-documented characteristic of modern multi-modal models, which excel at high-level semantic logic but inherently lack human sensitivity to subtle pixel artifacts. Ultimately, the substantial correlation in physical deduction and instruction execution definitively establishes our automated pipeline as a reliable, scalable, and physically grounded metric for PhyEditBench.
\section{Conclusion}\label{sec:conclusion}

In this paper, we introduced \textbf{PhyEditBench}, a pioneering high-resolution benchmark designed to rigorously evaluate the physics-based reasoning capabilities of instruction-guided image editing models. Unlike previous datasets that focus on static semantic modifications, our benchmark encompasses diverse real-world physical dynamics across fine-grained temporal stages, alongside challenging counterfactual anti-physics scenarios. Our comprehensive evaluation of current SOTA models exposed their critical limitations in understanding intuitive physics and dynamic state transitions. To bridge this gap, we proposed PhyWorld, a training-free framework that harnesses the temporal causality of pretrained video generation models as an implicit reasoning engine. By integrating test-time scaling with a latent reduction strategy, our method achieves comparable physical plausibility and visual consistency. Ultimately, we hope PhyEditBench will inspire future research toward equipping multi-modal generative models with a robust and dynamic understanding of the physical world.

\section*{Acknowledgements}
This work was supported in part by the National Natural Science Foundation of China, under Grant Nos. 62192783, 62276128, and 62406140; the Young Elite Scientists Sponsorship Program by China Association for Science and Technology, under Grant No. 2023QNRC001; the Key Research and Development Program of Jiangsu Province, under Grant No. BE2023019; and the Jiangsu Natural Science Foundation, under Grant Nos. BK20221441 and BK20241200. The authors would like to thank the support of Huawei Ascend Cloud Ecological Development Project.


%
%
\bibliographystyle{splncs04}
\bibliography{main}

\end{document}